%% file: main.tex
\documentclass[runningheads]{llncs}

\usepackage[T1]{fontenc}
\usepackage{microtype}
\usepackage{lmodern}

\usepackage{amsmath,amssymb}
\usepackage{mathtools}
\mathtoolsset{showonlyrefs=true}

\usepackage{array}
\usepackage{multirow}
\usepackage{booktabs}
\usepackage[table]{xcolor}

\usepackage{algorithm}
\usepackage{algpseudocode}

\usepackage{graphicx}
\usepackage{standalone}          
\usepackage{tikz}
\usetikzlibrary{arrows.meta,calc,positioning,fit,backgrounds}

\usepackage{hyperref}

\newlength{\normwidth}
\newcommand{\B}[1]{%
  \settowidth{\normwidth}{#1}
  \makebox[\normwidth][c]{\textbf{#1}}
}

\newcommand{\indicatorFunction}{\mathbf{1}}
\DeclareMathOperator*{\argmax}{argmax}
\newcommand{\ccdot}{\,\cdot\,}

\newcommand{\probsimplex}{\Delta^{K-1}}     

\newcommand{\entropy}{e}                      
\newcommand{\miaClf}{h_{\phi}}                

\newcommand{\kldiv}{\mathrm{KL}}
\newcommand{\jsdiv}{\mathrm{JS}}

\newcommand{\Minit}{M_{\mathrm{init}}}
\newcommand{\Mo}{M_{\mathrm{o}}}
\newcommand{\Mu}{M_{\mathrm{u}}}
\newcommand{\Mr}{M_{\mathrm{r}}}

\newcommand{\Dsettrain}{D}
\newcommand{\Dsetftrain}{D_{\mathrm{f}}}
\newcommand{\Dsetrtrain}{D_{\mathrm{r}}}
\newcommand{\Dsettest}{D'}
\newcommand{\Dsetftest}{D_{\mathrm{f}}'}
\newcommand{\Dsetrtest}{D_{\mathrm{r}}'}
\newcommand{\Dgen}{\tilde{D}}

\newcommand{\trainSeed}{s_{\mathrm{t}}}
\newcommand{\unlearnSeed}{s_{\mathrm{u}}}
\newcommand{\evalSeed}{s_{\mathrm{e}}}

\newcommand{\Layer}{\mathrm{Layer}}
\newcommand{\Activ}{\mathrm{Activ}}
\newcommand{\Compl}{\mathrm{Compl}}
\newcommand{\Acc}{\mathrm{Acc}}
\newcommand{\Loss}{\mathrm{Loss}}
\newcommand{\MIA}{\mathrm{MIA}}

\newcommand{\DAcc}{\Delta\Acc}
\newcommand{\DLoss}{\Delta\Loss}
\newcommand{\DMIA}{\Delta\MIA}

\newcommand{\Time}{\mathrm{Time}}
\newcommand{\Memory}{\mathrm{Memory}}
\newcommand{\Util}{\mathrm{Util}}
\newcommand{\tstart}{t_{\mathrm{start}}}
\newcommand{\tend}{t_{\mathrm{end}}}
\newcommand{\mbaseline}{m_{\mathrm{baseline}}}

\begin{document}

\title{SUPREME: A Multi-GPU Framework for Reproducible Image Unlearning Method Evaluation}
\titlerunning{SUPREME: A Framework for Image Unlearning Evaluation}
\author{Petros Andreou\inst{1} \and
Jamie Lanyon\inst{1} \and
Axel Finke\inst{2} \and
Georgina Cosma\inst{1}}
\authorrunning{P. Andreou et al.}
\institute{Department of Computer Science, School of Science,
Loughborough University, Loughborough, UK\\
\email{\{p.andreou2,j.lanyon,g.cosma\}@lboro.ac.uk}
\and
School of Mathematics, Statistics and Physics, Newcastle University,
Newcastle upon Tyne, NE1 7RU, UK\\
\email{axel.finke@newcastle.ac.uk}}
\maketitle

\begin{abstract}
Machine unlearning removes the influence of specific training data from a trained model without retraining it from scratch. Evaluating an unlearning method requires repeating training, unlearning, and evaluation across multiple seeds, which is computationally expensive. To our knowledge, existing image classification unlearning frameworks run on a single GPU, which limits how many seeds can be evaluated in reasonable time. We introduce SUPREME, an open-source framework that distributes these stages across multiple GPUs. SUPREME makes three contributions: a registry-based design for adding new methods, metrics, models, and scenarios; a multi-GPU architecture supporting multiple accelerators and precision modes; and a demonstration on Pins Face Recognition using ResNet18 and ViT under full-class and random-sample unlearning across ten seeds. The framework is available at \url{https://github.com/pedroandreou/supreme-unlearning}.
\end{abstract}

\section{Introduction}\label{sec:introduction}

Machine unlearning removes the influence of specific training data from a trained model without retraining it from scratch. Several approximate methods have been proposed, including fine-tuning, teacher--student distillation~\cite{chundawat2023can}, impair-and-repair schemes~\cite{tarun2023fast}, and parameter-importance methods~\cite{foster2024fast,foster2024loss}. Evaluating these methods requires metrics spanning multiple criteria: forgetting, utility, behavioural and parametric equivalence to a model retrained on the retain set, and privacy.

Three factors motivate reproducible evaluation of unlearning. First, the number of proposed methods and metrics is increasing, and comparing them under the same training setup and seeds is required to identify which methods perform best. Second, evaluation now covers larger model architectures such as Vision Transformers and larger datasets, which increases the compute cost of every training run. Third, recent work shows that single-seed results can misrepresent a method's performance, because outcomes depend on initial weights, data ordering, and the stochasticity of the unlearning step~\cite{cadet2025deep,lanyon2025limitation}. Repeating the full pipeline across multiple seeds is therefore required for a consistent comparison. 

Evaluation frameworks for image classification unlearning include MUBox~\cite{li2025mubox}, which benchmarks methods across several scenarios; ERASURE~\cite{dangelo2025erasure}, a modular toolkit for different data domains; and Deep Unlearn~\cite{cadet2025deep}, which tests methods across multiple random initialisations. Each runs on a single GPU at full precision, and of these only Deep Unlearn supports multi-seed evaluation.

We introduce SUPREME (Standardised Unlearning Platform for Reproducible Method Evaluation), an open-source framework for image classification unlearning. Our contributions are as follows. 

\begin{itemize}
\item \textbf{An extensible framework} with a registry-based design covering datasets, model architectures, unlearning methods, evaluation metrics, and unlearning scenarios. New components are added by implementing an interface and registering a module path, without modifying framework code.

\item \textbf{A hardware-agnostic architecture with multi-GPU support} built on PyTorch and Lightning Fabric, including DDP, FSDP, and DeepSpeed ZeRO. Distribution applies to training, unlearning, and evaluation --- to our knowledge the first image classification unlearning framework to do so.

\item \textbf{A demonstration on Pins Face Recognition.} We evaluate the integrated methods on ResNet18 and ViT under full-class and random-sample unlearning across ten seeds.
\end{itemize}

The remainder of this paper is organised as follows. Section~\ref{sec:supreme_framework} presents the SUPREME framework, Section~\ref{sec:experiments} details the experimental methodology, and Section~\ref{sec:results} reports and discusses the results. The appendices provide the component registry (Appendix~\ref{sec:appendix_components}), the metric definitions (Appendix~\ref{sec:unlearning_metric_definitions}), and additional results (Appendix~\ref{sec:appendix_raw_values}).

\section{SUPREME Framework}\label{sec:supreme_framework}

\subsection{Notation}\label{subsec:notation}

Let $\Minit$ be a model parameterised by initial weights which may be randomly initialised or pre-trained and let $\Mo$ be the model trained on some \emph{training} set $\Dsettrain$ consisting of image--label pairs $(x, y)$. Machine unlearning seeks to remove the influence of some \emph{forget} set $\Dsetftrain \subseteq \Dsettrain$ from $\Mo$. This is done by applying some unlearning method to $\Mo$ to obtain an \emph{unlearned} model $\Mu$. The goal is that $\Mu$ mimics as closely as possible the \emph{retrained} baseline $\Mr$, i.e., the model which has been trained from scratch on the \emph{retain} set $\Dsetrtrain \coloneqq \Dsettrain \setminus \Dsetftrain$. Unlearning performance is assessed by evaluating $\Mu$ on a \emph{test forget} set $\Dsetftest$ and a \emph{test retain} set $\Dsetrtest$ which we now specify based on some unseen dataset $\Dsettest$.

SUPREME supports two types of unlearning scenarios: targeted and random-sample unlearning. These differ in how the set of \emph{forget targets} $C$ is defined:

\begin{itemize}
    \item \emph{Targeted} unlearning removes a label-defined subset of the training data, i.e., each $c \in C$ is a class (or sub-class) label and the corresponding forget set is $\Dsetftrain \coloneqq \{(x, y) \in \Dsettrain \mid y = c\}$.  
   \item \emph{Random-sample} unlearning removes a random subset of the training samples, i.e., each $c \in C$ is a number in $(0, 1]$ and the corresponding forget set $\Dsetftrain$ is a subset -- sampled uniformly at random -- of $\Dsettrain$ of size $\lceil c |\Dsettrain| \rceil$.
\end{itemize}
The two scenarios also differ in how the forget and retain test sets, $\Dsetftest$ and $\Dsetrtest$, are defined: targeted unlearning uses held-out samples of the forgotten target ($\Dsetftest \coloneqq \{(x,y) \in \Dsettest \mid y = c\}$ and $\Dsetrtest \coloneqq \Dsettest \setminus \Dsetftest$) to test whether unlearning generalises beyond the training samples, while random-sample unlearning has no such held-out set and evaluates on the training forget samples directly ($\Dsetftest \coloneqq \Dsetftrain$ and $\Dsetrtest \coloneqq \Dsettest$).

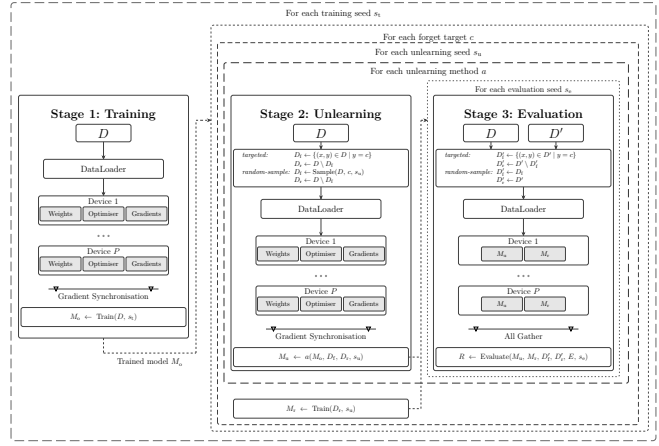
\begin{figure}[t]
    \centering
    \resizebox{0.7\textwidth}{!}{\input{supreme_diagram}} \caption{SUPREME's seeded multi-stage pipeline. All three stages run across $P$ devices. \emph{Gradient Synchronisation} (Stages~1--2) averages gradients across devices after each backward pass so all devices apply the same parameter update. \emph{Result Aggregation} (Stage~3), implemented as an all-gather, collects per-batch metric values from all devices and averages them into the final metric value. 
    The training forget/retain split is computed once per unlearning seed and the test split once per method, not per inner-loop iteration.}
    \label{fig:supreme_pipeline}
\end{figure}

\subsection{Pipeline}\label{subsec:pipeline}
Figure~\ref{fig:supreme_pipeline} shows the SUPREME pipeline and Algorithm~\ref{alg:supreme} provides details. Stage~1 trains $\Mo$ on $\Dsettrain$. For each forget target $c \in C$, the framework trains $\Mr$ on $\Dsetrtrain$. For each unlearning method $a \in A$, Stage~2 applies $a$ to $\Mo$ to obtain $\Mu$, and Stage~3 evaluates $\Mu$ against $\Mr$ using the configured metrics. The pipeline has three properties: seeded randomness, distributed execution, and registry-based extensibility.

\textbf{Seeded randomness.} Each seed determines initial weights, data ordering, the random forget-set draw (under the random-sample scenario), and the stochastic operations inside the unlearning step. Running every method under the same seed configuration ensures that every $a \in A$ is evaluated under identical starting conditions for a given seed, isolating method differences from pipeline randomness. The framework allows independent seeds for the training, unlearning, and evaluation stages, enabling users to isolate the effect of each stage when needed.

\textbf{Distributed execution.} All three stages execute across $P$ devices. Training and unlearning use gradient synchronisation, where gradients are averaged across devices after each backward pass so all devices apply the same parameter update. Evaluation uses result aggregation, where per-batch metric values are collected across devices and averaged into the final metric value. Distribution applies to every stage, including the retrained baseline, which dominates total runtime when $|C|$ or the number of seeds is large.

\textbf{Registry-based extensibility.} Datasets, model architectures, unlearning methods, evaluation metrics, and unlearning scenarios are registered via module paths. New components are added by implementing the required interface and registering the module path, without modifying framework code.

\begin{algorithm}[htb!]
\caption{SUPREME: Seeded Multi-Stage Pipeline}
\label{alg:supreme}
\begin{algorithmic}[1]
\Require Training dataset $\Dsettrain$, Test dataset $\Dsettest$, Unlearning methods $A$, Metrics $E$, number of training seeds $I \in \mathbb{N}$, number of unlearning seeds per training seed $J \in \mathbb{N}$, number of evaluation seeds per unlearning seed $K \in \mathbb{N}$, 
Forget targets $C$, Devices $P$, Scenario type $\tau \in \{\text{random-sample}, \text{targeted}\}$
\Ensure Results dictionary $R$ for each tuple $(i, j, k, c, a)$
\For{$i \in \{1,\dotsc,I\}$} 
    \State $\trainSeed \gets i$ \Comment{Training seed}
    \State \textbf{Stage 1: Training} \Comment{Distributed across $P$}
    \State $\Mo \gets \Call{Train}{\Dsettrain, \trainSeed}$ \Comment{Train from initial parameters using Seed $\trainSeed$}
    \For{each forget target $c \in C$}
        \For{$j \in \{1,\dotsc, J\}$} 
        \State $\unlearnSeed \gets (i-1)J + j$ \Comment{Unlearning seed}
            \If{$\tau = \text{targeted}$}
                    \State $\Dsetftrain \gets \{(x,y) \in \Dsettrain \mid y = c\}$
            \ElsIf{$\tau = \text{random-sample}$}
                    \State $\Dsetftrain  \gets \Call{Sample}{\Dsettrain, c, \unlearnSeed}$ \Comment{Sample $c \cdot 100\,\%$ of $\Dsettrain$ using Seed $\unlearnSeed$}
            \EndIf
            \State $\Dsetrtrain \gets \Dsettrain \setminus \Dsetftrain$
            \State \textbf{Stage 2: Unlearning} \Comment{Distributed across $P$}
                \State $\Mr \gets \Call{Train}{\Dsetrtrain, \unlearnSeed}$ \Comment{Retrained baseline using Seed $\unlearnSeed$}
                \For{each method $a \in A$}
                \State $\Mu \gets a(\Mo, \Dsetftrain, \Dsetrtrain, \unlearnSeed)$ \Comment{Apply unlearning using Seed $\unlearnSeed$}
                \State \textbf{Stage 3: Evaluation} \Comment{Distributed across $P$}
                     \If{$\tau = \text{targeted}$}
                    \State $\Dsetftest \gets \{(x,y) \in \Dsettest \mid y = c\}$; $\Dsetrtest \gets \Dsettest \setminus \Dsetftest$
                \ElsIf{$\tau = \text{random-sample}$}
                    \State $\Dsetftest \gets \Dsetftrain$; $\Dsetrtest \gets \Dsettest$
                \EndIf
                \For{$k \in \{1,\dotsc, K\}$} 
                \State $\evalSeed \gets (i-1) J K + (j-1) K + k$ \Comment{Evaluation seed}
                \State $R[(i,j,k,c,a)] \gets \Call{Evaluate}{\Mu, \Mr, \Dsetftest, \Dsetrtest, E, \evalSeed}$
                \EndFor
            \EndFor
        \EndFor
    \EndFor
\EndFor
\end{algorithmic}
\end{algorithm}

\subsection{Implementation}\label{subsec:framework_implementation}

SUPREME is built on PyTorch and Lightning Fabric. The framework supports DDP, FSDP, and DeepSpeed ZeRO-1/2/3 as distributed strategies, selectable at runtime. Two optimisations reduce per-run overhead when the pipeline is repeated across many seeds: enabling tensor cores for matrix operations, and accelerated model initialisation that defers parameter materialisation until the model is moved to its device.

\textbf{Vision models.} SUPREME registers two model architectures. ResNet18~\cite{he2016deep} is a four-block residual convolutional network with 64--512 channels per block, trained from scratch on $32 \times 32$ inputs. ViT~\cite{dosovitskiy2021image} is fine-tuned from \texttt{google/vit-base-patch16-224}, which divides $224 \times 224$ inputs into $16 \times 16$ patches and processes them through 12 transformer layers with 12 attention heads and 768 hidden dimensions. Depending on the selected distributed strategy, the weights, optimiser state, and gradients of these models (shown in Stages~1--2 of Figure~\ref{fig:supreme_pipeline}) are replicated across the $P$ devices (DDP) or partitioned among them (FSDP and DeepSpeed ZeRO).

\textbf{Image preprocessing.} ResNet18 inputs are resized to $32 \times 32$ with standard training-time augmentation (random crop, horizontal flip, rotation) and per-channel normalisation computed from the training set. ViT inputs are resized to $256 \times 256$, centre-cropped to $224 \times 224$, horizontally flipped, and normalised with ImageNet statistics, as required by the pre-trained model. Augmentation is disabled during unlearning and evaluation.

\textbf{Evaluation metrics.} Unlearning quality is evaluated under seven criteria: behavioural equivalence, parametric equivalence, privacy, forgetting, utility, efficiency, and resources. Definitions and the metrics under each criterion are in Table~\ref{tab:metrics_summary} in Appendix~\ref{sec:unlearning_metric_definitions}.

\section{Experimental Methodology}\label{sec:experiments}
We demonstrate SUPREME on Pins Face Recognition~\cite{burak2020pins}, an image classification benchmark of 17{,}534 facial images across 105 celebrity identities. The preprocessing follows Section~\ref{subsec:framework_implementation}. ResNet18 uses statistics computed from the training set, $\mu = (0.516, 0.419, 0.373)$ and $\sigma = (0.286, 0.255, 0.246)$. ViT uses ImageNet statistics, $\mu = (0.485, 0.456, 0.406)$ and $\sigma = (0.229, 0.224, 0.225)$. We evaluate two unlearning scenarios. Full-class unlearning removes all samples for five identities: alex\_lawther, bill\_gates, danielle\_panabaker, hugh\_jackman, and josh\_radnor. Random-sample unlearning removes a 0.1\,\% subset of training samples drawn from across all classes. We use $I = 10$ training runs with seeds 260--269 (the training-seed index $i$ in Algorithm~\ref{alg:supreme} maps to these values), with a single unlearning and evaluation seed per training seed ($J = K = 1$). Varying the training seed surfaces the across-seed variance reported in Section~\ref{sec:results}; the separate training, unlearning, and evaluation seeds ($\trainSeed$, $\unlearnSeed$, $\evalSeed$) that Algorithm~\ref{alg:supreme} also supports would decompose that variance further but are not needed for this demonstration. At $J = K = 1$, the seed formulae in Algorithm~\ref{alg:supreme} reduce to $\unlearnSeed = \evalSeed = \trainSeed$. The unlearning methods listed in Table~\ref{tab:components} in Appendix~\ref{sec:appendix_components} are applied to $\Mo$, with $\Mr$ as the baseline. The experiments were run on a single NVIDIA L40S GPU (48~GB VRAM) to maintain exact numerical parity with the reference implementations~\cite{chundawat2023can,foster2024fast,foster2024loss}, following the same choice made by OpenUnlearning~\cite{dorna2025openunlearning} in the text domain.

\begin{table}[t]
\caption{Accuracy differences ($\DAcc$) and layer-wise distances ($\Layer$) between $\Mu$ and $\Mr$ on Pins Face Recognition (closer to 0 is better). Bold marks the best (closest to 0) value in each column within a model--scenario block. $\Layer$ is a weight-space metric with no dataset split. Full-class values average over 5 forget classes; random uses a 0.1\,\% forget set. Mean $\pm$ std across 10 seeds. Raw accuracy values are in Table~\ref{tab:pins_raw_values} in Appendix~\ref{sec:appendix_raw_values}. UNSIR is excluded from the random scenario by design~\cite{tarun2023fast}.}
\label{tab:pins_results_main}
\centering
\setlength{\tabcolsep}{5pt}
\begin{tabular}{@{}lll r@{.}l @{\,$\pm$\,} r@{.}l @{\hspace{1.5em}} r@{.}l @{\,$\pm$\,} r@{.}l @{\hspace{1.5em}} r@{.}l @{\,$\pm$\,} r@{.}l@{}}
\toprule
\textbf{Model} & \textbf{Scenario} & \textbf{Method} 
& \multicolumn{4}{c}{\textbf{$\DAcc_{\Dsetftest}$}} 
& \multicolumn{4}{c}{\textbf{$\DAcc_{\Dsetrtest}$}} 
& \multicolumn{4}{c}{\textbf{$\Layer$}} \\
\midrule
  \multirow{11}{*}{ResNet18} & \multirow{6}{*}{\small Full-class} & FT & 29&22 & 5&21 & 2&52 & 0&11 & \B{31}&\B{52} & \B{0}&\B{22} \\
   &  & BadT & 0&26 & 0&17 & \B{-2}&\B{35} & \B{0}&\B{27} & 31&72 & 0&34 \\
   &  & UNSIR & 89&44 & 1&98 & 2&51 & 0&12 & 32&25 & 0&34 \\
   &  & RL & \B{0}&\B{00} & \B{0}&\B{00} & 2&58 & 0&12 & 31&98 & 0&34 \\
   &  & SSD & 1&97 & 6&22 & -9&14 & 7&43 & 31&55 & 0&36 \\
   &  & LFSSD & \B{0}&\B{00} & \B{0}&\B{00} & -3&66 & 1&53 & 31&56 & 0&34 \\
\cmidrule{2-15}
   & \multirow{5}{*}{\small Random} & FT & \B{2}&\B{78} & \B{24}&\B{32} & \B{-2}&\B{60} & \B{15}&\B{25} & \B{37}&\B{70} & \B{7}&\B{91} \\
   &  & BadT & -35&00 & 25&93 & -32&47 & 29&83 & 41&99 & 8&78 \\
   &  & RL & -48&89 & 34&03 & -4&59 & 20&12 & 42&11 & 8&72 \\
   &  & SSD & -75&00 & 19&47 & -79&97 & 23&79 & 40&25 & 9&06 \\
   &  & LFSSD & -68&33 & 15&28 & -58&61 & 19&20 & 40&82 & 8&72 \\
\midrule
  \multirow{11}{*}{ViT} & \multirow{6}{*}{\small Full-class} & FT & 0&03 & 0&06 & -22&66 & 23&47 & 105&29 & 1&31 \\
   &  & BadT & 17&90 & 5&18 & -0&88 & 0&10 & \B{33}&\B{97} & \B{0}&\B{12} \\
   &  & UNSIR & 35&52 & 4&92 & -0&31 & 0&13 & 39&38 & 0&18 \\
   &  & RL & \B{0}&\B{00} & \B{0}&\B{00} & \B{0}&\B{20} & \B{0}&\B{05} & 36&92 & 0&18 \\
   &  & SSD & \B{0}&\B{00} & \B{0}&\B{00} & -2&40 & 0&99 & 63&38 & 2&58 \\
   &  & LFSSD & \B{0}&\B{00} & \B{0}&\B{00} & -3&88 & 2&11 & 86&77 & 6&07 \\
\cmidrule{2-15}
   & \multirow{5}{*}{\small Random} & FT & \B{8}&\B{33} & \B{5}&\B{40} & 1&42 & 0&12 & 60&40 & 0&19 \\
   &  & BadT & \B{-8}&\B{33} & \B{7}&\B{97} & \B{0}&\B{21} & \B{0}&\B{52} & \B{32}&\B{76} & \B{0}&\B{18} \\
   &  & RL & -76&67 & 11&94 & 1&33 & 0&12 & 35&71 & 0&20 \\
   &  & SSD & -55&00 & 37&99 & -54&09 & 45&60 & 171&80 & 84&31 \\
   &  & LFSSD & -89&44 & 6&11 & -90&70 & 6&59 & 231&91 & 15&22 \\
\bottomrule
\end{tabular}
\end{table}

\section{Results and Discussion}\label{sec:results}

Table~\ref{tab:pins_results_main} reports the test-accuracy differences between $\Mu$ and $\Mr$ on $\Dsetftest$ ($\DAcc_{\Dsetftest}$) and $\Dsetrtest$ ($\DAcc_{\Dsetrtest}$), and the layer-wise weight distance $\Layer$ between the two models. Table~\ref{tab:pins_results_appendix} in Appendix~\ref{sec:appendix_raw_values} reports activation distances and membership inference attack (MIA) score differences ($\DMIA$). Together these cover four of the seven evaluation criteria in Table~\ref{tab:metrics_summary} in Appendix~\ref{sec:unlearning_metric_definitions}: forgetting, utility, parametric equivalence, and privacy. 

\textbf{Across-seed variance.} Several method-and-scenario combinations produced large standard deviations over the ten seeds. The two largest forget-set accuracy ($\DAcc_{\Dsetftest}$) standard deviations both occurred in the random-sample scenario. Random Labels on ResNet18 gave $\DAcc_{\Dsetftest} = -48.89 \pm 34.03$ percentage points, and Selective Synaptic Dampening on ViT gave $-55.00 \pm 37.99$. A single seed can therefore differ from the ten-seed mean by tens of percentage points. Because only the training seed is varied ($J = K = 1$; see Section~\ref{sec:experiments}), the observed spread combines four sources of randomness: initial weights, data ordering, the random forget-set draw, and stochastic operations inside the unlearning method.

\textbf{Multi-criterion reporting.} In the full-class scenario, the mean MIA difference ($\DMIA$) in Table~\ref{tab:pins_results_appendix} stays within $\pm 0.05$ for every method on both model architectures, and seven of the twelve satisfy $|\DMIA| \le 0.02$. On the same runs, $\DAcc_{\Dsetftest}$ ranges from $0.00$ to $89.44$ percentage points on ResNet18 and from $0.00$ to $35.52$ on ViT. $\DMIA$ therefore does not separate the methods in this setting, whereas $\DAcc_{\Dsetftest}$ does. The methods may have similar privacy behaviour, or the MIA probe may be too weak on Pins Face Recognition.

\section{Conclusion}\label{sec:conclusion}
We introduced SUPREME, an open-source, registry-based framework that distributes training, unlearning, and evaluation across multiple GPUs, and demonstrated it on Pins Face Recognition across ten seeds. We verified that single- and multi-GPU execution yield the same evaluation-metric results; the numbers reported here were produced on a single GPU, matching the single-device setup of the integrated methods' reference implementations. The demonstration is confined to a single dataset, two model architectures, and two scenarios, so the reported numbers characterise the framework
rather than rank the methods. Future work will present these experiments on large-scale datasets, where distributing the pipeline across multiple GPUs is most beneficial, extend the demonstration to the other registered components, and use the per-stage seeds to decompose the across-seed spread into its training, unlearning, and evaluation parts.

\begin{credits}
\subsubsection{\ackname}
Petros Andreou is supported by a PhD studentship funded by Darktrace.
Darktrace had no role in the study design, data collection, analysis,
interpretation, or in the decision to submit this work for publication.

\end{credits}

\bibliographystyle{splncs04}
\bibliography{references}

\newpage
\appendix
\renewcommand{\theHsection}{appendix.\thesection}

\section{SUPREME Component Registry}
\label{sec:appendix_components}

\begin{table}[H]
\caption{Components supported by SUPREME. Each item in the right column is a registered implementation that can be selected at runtime. Datasets, models, unlearning methods, evaluation metrics, and unlearning scenarios are registry-based and can be extended by implementing the relevant interface and registering the module path. The remaining components (accelerators, precision modes, and distributed strategies) are provided via Lightning Fabric; the listed options represent the full spectrum of supported hardware and execution configurations.}
\label{tab:components}
\centering
\small
\begin{tabular}{@{}>{\raggedright\arraybackslash}p{3cm}@{\hspace{1.2cm}}>{\raggedright\arraybackslash}p{8cm}@{}}
\toprule
\textbf{Component} & \textbf{Supported Implementations} \\
\midrule
\multicolumn{2}{c}{\cellcolor{gray!15}\textit{Registry-based components (user-extensible)}} \\
Datasets
  & CIFAR-10, CIFAR-20, CIFAR-100, PinsFaceRecognition, Caltech-101 \\[4pt]
Models
  & ResNet18, Vision Transformer (ViT) \\[4pt]
Unlearning Methods
  & Retrain, Fine-Tuning (FT), Bad Teacher (BadT), Random Labels (RL), Unlearning by Selective Impair and Repair (UNSIR), Selective Synaptic Dampening (SSD), Loss-Free Selective Synaptic Dampening (LFSSD) \\[4pt]
Evaluation Metrics
  & see Table~\ref{tab:metrics_summary} in Appendix~\ref{sec:unlearning_metric_definitions} \\[4pt]
Unlearning Scenarios
  & Targeted (full-class, sub-class), Random sample \\[4pt]
\midrule
\multicolumn{2}{c}{\cellcolor{gray!15}\textit{Provided via Lightning Fabric}} \\
Accelerators
  & CPU, GPU, MPS, TPU \\[4pt]
Precision Modes
  & 64-bit Double Precision (64-true), 32-bit Full Precision (32-true), 16-bit Mixed Precision (16-mixed), 16-bit Brain Floating Point Mixed Precision (bf16-mixed), 16-bit True Precision (16-true), 16-bit True Brain Floating Point Precision (bf16-true), 8-bit Floating Point via NVIDIA TransformerEngine with bfloat16 weights (transformer-engine), 8-bit Floating Point via NVIDIA TransformerEngine with float16 weights (transformer-engine-float16), 8-bit Integer Inference via BitsandBytes (int8), 8-bit Integer Fine-Tuning via BitsandBytes (int8-training), 4-bit NormalFloat via BitsandBytes (nf4), 4-bit NormalFloat with Double Quantisation via BitsandBytes (nf4-dq), 4-bit Floating Point via BitsandBytes (fp4), 4-bit Floating Point with Double Quantisation via BitsandBytes (fp4-dq) \\[4pt]
Distributed Strategies
  & Distributed Data Parallel (DDP), Fully Sharded Data Parallel (FSDP), Zero Redundancy Optimiser (DeepSpeed ZeRO) Stage-1/2/3 \\[4pt]
\bottomrule
\end{tabular}
\end{table}

\clearpage
\newpage

\section{Unlearning Metrics}\label{sec:unlearning_metric_definitions} 
The following metrics evaluate machine unlearning quality by comparing the unlearned model $\Mu$ against the retrained model $\Mr$. In the following definitions, $M$ denotes some model (e.g., the unlearned or retrained model). Throughout, $M(x) \in \probsimplex$ denotes the softmax probability vector produced by $M$ on input $x$, where $K$ is the number of classes and $\probsimplex$ is the $(K-1)$-simplex. $\Dgen$ denotes some generic dataset.

\subsection{Model evaluation metrics}\label{subsec:model_evaluation_metrics}

\textbf{Layer-wise distance.} Layer-wise distance quantifies parameter changes during unlearning by measuring the Euclidean distance between the weights of $\Mu$ and $\Mr$:
\begin{equation}\label{eq:layer_distance}
  \Layer \coloneqq \sqrt{\sum_{l=1}^L \sum_{i=1}^{n_l} (w_{\mathrm{u}}^{l,i} - w_{\mathrm{r}}^{l,i})^2} \in [0, +\infty),
\end{equation}
where $w_{\mathrm{u}}^{l,i}$ and $w_{\mathrm{r}}^{l,i}$ represent the $i$th weight parameter in layer $l$ of models $\Mu$ and $\Mr$ respectively, $n_l$ denotes the number of learnable parameters in layer $l$, and $L$ is the total number of layers. Smaller distance indicates better unlearning (zero means $\Mu$ and $\Mr$ have identical weights); larger distance indicates worse unlearning.

\textbf{Activation distance.} Activation distance quantifies the difference between the final-layer activations of $\Mu$ and $\Mr$ on $\Dgen$. Implementations include the L1-norm between softmax outputs \cite{golatkar2020forgetting} and the L2-norm \cite{chundawat2023can}; this work uses L2-norm:
\begin{equation}\label{eq:actv_dist}
\Activ_{\Dgen} \coloneqq \sqrt{\frac{1}{|\Dgen|} \sum_{(x,y) \in \Dgen} \|\Mr(x) - \Mu(x)\|_2^2} \in [0, \sqrt{2}],
\end{equation}
where $\|\cdot\|_2$ is the Euclidean norm. Smaller activation distance indicates better unlearning (zero means $\Mu$ and $\Mr$ produce identical outputs on samples in $\Dgen$); larger distance indicates worse unlearning.

\textbf{JS-Divergence.} Jensen--Shannon divergence symmetrically quantifies similarity between probability distributions, derived from the Kullback--Leibler divergence. It compares output distributions of $\Mr$ and $\Mu$ on $\Dgen$:
\begin{align}\label{eq:jsdivform}
 \jsdiv_{\Dgen} \coloneqq \frac{1}{2|\Dgen|} \sum_{(x',y') \in \Dgen} \Bigl[ \kldiv(\Mr(x') \parallel m(x'))  + \kldiv(\Mu(x') \parallel m(x')) \Bigr] \in [0, 1],
\end{align}
where $m(x') \coloneqq (\Mr(x') + \Mu(x'))/2$ is the average output of both models for sample $x'$ in $\Dgen$, and $\kldiv(\ccdot\!\parallel\!\ccdot)$ is the Kullback--Leibler divergence. Lower JS-divergence indicates better unlearning (zero means $\Mu$ and $\Mr$ produce identical output distributions on samples in $\Dgen$; one means completely different distributions).

\textbf{Completeness.} Completeness quantifies prediction consistency between $\Mr$ and $\Mu$ on $\Dgen$. Predictions are considered complete when both models agree, regardless of correctness, focusing on consistency rather than accuracy:
\begin{align}\label{eq:completeness_perc}
  \Compl_{\Dgen} \coloneqq \frac{1}{|\Dgen|} \sum_{(x, y) \in \Dgen} \indicatorFunction\Bigl\{ \argmax_k [\Mr(x)]_k = \argmax_k [\Mu(x)]_k \Bigr\} \in [0, 1],
\end{align}
where $[z]_k$ is the $k$th element of a vector $z$, and $\argmax_k$ returns the class with the highest predicted probability. Lower completeness indicates worse unlearning (approaching zero means $\Mu$ and $\Mr$ produce completely different predictions on samples in $\Dgen$; one means identical predictions on all samples).

\textbf{Accuracy.} Accuracy differences ($\DAcc_{\Dgen}$) computes the difference in predictive performance between $\Mu$ and $\Mr$ on $\Dgen$ \cite{golatkar2020forgetting}:
\begin{equation}\label{eq:accuracy}
\DAcc_{\Dgen}\coloneqq \Acc_{\Dgen}(\Mu) - \Acc_{\Dgen}(\Mr)\in [-1, 1],
\end{equation}
where
\begin{equation}\label{eq:accuracy_def}
\Acc_{\Dgen}(M) = \frac{1}{|\Dgen|} \sum_{(x,y) \in \Dgen} \indicatorFunction\{ \argmax_k [M(x)]_k = y \}\in [0, 1].
\end{equation}
When $M = \Mu$ is the unlearned model, we suppress the argument and write $\Acc_{\Dgen} \coloneqq \Acc_{\Dgen}(\Mu)$.
$\DAcc_{\Dgen}$ closer to zero indicates better unlearning (zero means $\Mu$ and $\Mr$ have identical accuracy on $\Dgen$); larger $|\DAcc_{\Dgen}|$ indicates worse unlearning.

\textbf{Loss.} Loss differences ($\DLoss_{\Dgen}$) computes the difference in cross-entropy error between $\Mu$ and $\Mr$ on $\Dgen$:
\begin{equation}\label{eq:loss}
\DLoss_{\Dgen} = \Loss_{\Dgen}(\Mu) - \Loss_{\Dgen}(\Mr) \in \mathbb{R},
\end{equation}
where
\begin{equation}\label{eq:loss_def}
\Loss_{\Dgen}(M) = -\frac{1}{|\Dgen|} \sum_{(x,y) \in \Dgen} \log [M(x)]_y\in [0, +\infty).
\end{equation}
When $M = \Mu$ is the unlearned model, we suppress the argument and write $\Loss_{\Dgen} \coloneqq \Loss_{\Dgen}(\Mu)$.
$\DLoss_{\Dgen}$ closer to zero indicates better unlearning (zero means $\Mu$ and $\Mr$ have identical loss on $\Dgen$); larger $|\DLoss_{\Dgen}|$ indicates worse unlearning.

\textbf{Membership Inference Attack.} Membership inference attack differences ($\DMIA$) measures the difference in data leakage risk between $\Mu$ and $\Mr$ on the forget training set, determining whether a data point was in the model's training set $\Dsettrain$. A logistic regression classifier trained on the entropy of softmax probabilities distinguishes $\Dsetrtrain$ (labelled as 1) from $\Dsettest$ (labelled as 0), then evaluates $\Dsetftrain$ to measure retained information \cite{chundawat2023can}:
\begin{align}\label{eq:mia}
 \DMIA \coloneqq \MIA(\Mu)
- \MIA(\Mr)\in [-1, 1],
\end{align}
where
\begin{equation}\label{eq:mia_def}
\MIA(M) = \frac{1}{|\Dsetftrain|} \sum_{(x,y) \in \Dsetftrain} \indicatorFunction\{ \miaClf(\entropy(M(x))) = 1 \}\in [0, 1].
\end{equation}
When $M = \Mu$ is the unlearned model, we suppress the argument and write $\MIA \coloneqq \MIA(\Mu)$.
Above, $\indicatorFunction$ is the indicator function, $\miaClf$ is the logistic regression classifier with parameters $\phi$, trained on $\{(\entropy(M(x)), 1) : (x,y) \in \Dsetrtrain\}$ (retain samples labelled as 1) and $\{(\entropy(M(x)), 0) : (x,y) \in \Dsettest\}$ (test samples labelled as 0), and $\entropy(\cdot)$ is the entropy function applied to the model output probabilities:
\begin{equation}\label{eq:entropy_func}
\entropy(p) = -\sum_{k=1}^{K} p_k \log p_k\in [0, \log K],
\end{equation}
where $p = (p_1, \ldots, p_K) \in \probsimplex$. Values of $\DMIA$ closer to zero indicates better unlearning (zero means $\Mu$ and $\Mr$ have identical membership inference vulnerability); larger $|\DMIA|$ indicates worse unlearning.

\subsection{Computational Cost Metrics}\label{subsec:computational_cost_metrics}
The following metrics measure the computational cost of the unlearning procedure rather than the quality of the resulting model. They capture the practical viability and scalability of unlearning methods, measuring time and resource consumption during the execution of the unlearning method.

\textbf{Time.} Time measures the elapsed wall-clock duration of the unlearning procedure:
\begin{equation}\label{eq:time_def}
\Time \coloneqq \tend - \tstart \in [0, +\infty),
\end{equation}
where $\tstart$ is the timestamp recorded immediately before the unlearning method begins and $\tend$ is the timestamp recorded immediately after it terminates, both in seconds. Lower time indicates better efficiency; higher time indicates worse efficiency.

\textbf{Memory Usage.} Memory usage captures the peak memory consumed by the process during unlearning on the active accelerator (see Table~\ref{tab:components})~\cite{xu2024machine}:
\begin{equation}\label{eq:mem_def}
\Memory \coloneqq \max_{t \in [0, T]} \frac{m_{t} - \mbaseline}{1024^3} \in [0, +\infty),
\end{equation}
where $T \coloneqq \Time$ is the elapsed time defined in Eq.~\eqref{eq:time_def}, $t \in [0, T]$ is a time point during the procedure, $m_{t}$ is the amount of accelerator memory in bytes in use by the process at time $t$ (storing, e.g., model weights, activations, gradients, and any other process-resident tensors), $\mbaseline$ is the value of $m_{t}$ immediately before the procedure begins (subtracted so the metric isolates the procedure's contribution from any pre-existing footprint such as the runtime and imported libraries), and division by $1024^3$ converts bytes to gigabytes. Lower memory indicates better efficiency; higher memory indicates worse efficiency.

\textbf{Compute Utilisation.} Compute utilisation is the time-averaged percentage of accelerator compute resources used by the process during unlearning on the active accelerator (see Table~\ref{tab:components})~\cite{xu2024machine}:
\begin{equation}\label{eq:util_def}
\Util \coloneqq \frac{1}{N} \sum_{i=1}^{N} u_i \in [0, 100],
\end{equation}
where $u_i \in [0, 100]$ is the percentage of accelerator compute used by the process at the $i$-th measurement, and $N = fT$ is the number of measurements collected at sampling rate $f$\,Hz over duration $T$ seconds defined in Eq.~\eqref{eq:time_def} (a typical default is $f = 10$\,Hz). Lower utilisation indicates better efficiency; higher utilisation indicates worse efficiency.

\begin{table}[!hbt]
\caption{Evaluation metrics organised by seven criteria, each corresponding to a research question shown in grey.}
\label{tab:metrics_summary}
\centering
\setlength{\tabcolsep}{8pt}
\renewcommand{\arraystretch}{1.15}
\begin{tabular}{>{\raggedright\arraybackslash}p{3.2cm}>{\raggedright\arraybackslash}p{7cm}}
\toprule
\textbf{Criterion} & \textbf{Evaluation Metric} \\
\midrule
\multicolumn{2}{l}{\cellcolor{gray!15}\textit{Do $\Mu$ and $\Mr$ behave the same on arbitrary test inputs?}} \\
\multirow{3}{3.2cm}{Behavioural Equivalence} & Whole-set Activation Distance ($\Activ_{\Dsettest}$) \\
 & Whole-set JS-Divergence ($\jsdiv_{\Dsettest}$) \\
 & Whole-set Completeness ($\Compl_{\Dsettest}$) \\
\midrule
\multicolumn{2}{l}{\cellcolor{gray!15}\textit{Do $\Mu$ and $\Mr$ share the same internal parameters?}} \\
Parametric Equivalence & Layer-wise Distance ($\Layer$) \\
\midrule
\multicolumn{2}{l}{\cellcolor{gray!15}\textit{Which method best prevents information leakage about forgotten data?}} \\
Privacy & Membership Inference Attack ($\DMIA$) \\
\midrule
\multicolumn{2}{l}{\cellcolor{gray!15}\textit{Which method most effectively removes forget-set knowledge?}} \\
\multirow{5}{*}{Forgetting} & Forget-set Accuracy ($\DAcc_{\Dsetftest}$) \\
 & Forget-set Loss ($\DLoss_{\Dsetftest}$) \\
 & Forget-set Completeness ($\Compl_{\Dsetftest}$) \\
 & Forget-set Activation Distance ($\Activ_{\Dsetftest}$) \\
 & Forget-set JS-Divergence ($\jsdiv_{\Dsetftest}$) \\
\midrule
\multicolumn{2}{l}{\cellcolor{gray!15}\textit{Which method best preserves performance on the retain set?}} \\
\multirow{7}{*}{Utility} & Retain-set Accuracy ($\DAcc_{\Dsetrtest}$) \\
 & Retain-set Loss ($\DLoss_{\Dsetrtest}$) \\
 & Retain-set Completeness ($\Compl_{\Dsetrtest}$) \\
 & Retain-set Activation Distance ($\Activ_{\Dsetrtest}$) \\
 & Retain-set JS-Divergence ($\jsdiv_{\Dsetrtest}$) \\
 & Whole-set Accuracy ($\DAcc_{\Dsettest}$) \\
 & Whole-set Loss ($\DLoss_{\Dsettest}$) \\
\midrule
\multicolumn{2}{l}{\cellcolor{gray!15}\textit{Which method is fastest?}} \\
Efficiency & Time ($\Time$) \\
\midrule
\multicolumn{2}{l}{\cellcolor{gray!15}\textit{Which method uses the least computational resources?}} \\
\multirow{2}{*}{Resources} & Memory Usage ($\Memory$) \\
 & Compute Utilisation ($\Util$) \\
\bottomrule
\end{tabular}
\end{table}

\clearpage
\newpage

\section{Pins face additional results}
\label{sec:appendix_raw_values}

\begin{table}[H]
\caption{MIA differences ($\DMIA$) and Activation distances ($\Activ_{\Dgen}$) for full-class and random unlearning on Pins Face Recognition. $\DMIA$ is the difference between the unlearned model $\Mu$ and retrained model $\Mr$ (closer to 0 is better). $\Activ$ represents raw distances between the unlearned model $\Mu$ and retrained model $\Mr$ (lower is better). Full-class values are averaged over the 5 forget classes per seed; random values are for the 0.1\,\% forget percentage. All values are mean $\pm$ std across 10 seeds. Raw $\MIA$ values for $\Mu$ and $\Mr$ are reported in Table~\ref{tab:pins_raw_values}. UNSIR is excluded from the random scenario by design.}
\label{tab:pins_results_appendix}
\centering
\setlength{\tabcolsep}{5pt}
\begin{tabular}{@{}lll r@{.}l @{\,$\pm$\,} r@{.}l @{\hspace{1.5em}} r@{.}l @{\,$\pm$\,} r@{.}l @{\hspace{1.5em}} r@{.}l @{\,$\pm$\,} r@{.}l@{}}
\toprule
\textbf{Model} & \textbf{Scenario} & \textbf{Method} & \multicolumn{4}{c}{\textbf{$\Activ$ on $\Dsetftest$}} & \multicolumn{4}{c}{\textbf{$\Activ$ on $\Dsetrtest$}} & \multicolumn{4}{c}{\textbf{$\DMIA$}} \\
\midrule
  \multirow{11}{*}{ResNet18} & \multirow{6}{*}{\small Full-class} & FT & 0&68 & 0&01 & 1&04 & 0&00 & -0&05 & 0&01 \\
   &  & BadT & \B{0}&\B{49} & \B{0}&\B{01} & 1&00 & 0&00 & -0&05 & 0&01 \\
   &  & UNSIR & 0&96 & 0&02 & 1&04 & 0&00 & \B{-0}&\B{00} & \B{0}&\B{03} \\
   &  & RL & 0&99 & 0&01 & 1&05 & 0&00 & 0&04 & 0&02 \\
   &  & SSD & 0&63 & 0&04 & \B{0}&\B{97} & \B{0}&\B{02} & 0&01 & 0&07 \\
   &  & LFSSD & 0&60 & 0&01 & \B{0}&\B{97} & \B{0}&\B{01} & -0&03 & 0&01 \\
\cmidrule{2-15}
   & \multirow{5}{*}{\small Random} & FT & \B{0}&\B{33} & \B{0}&\B{11} & 0&98 & 0&15 & -0&06 & 0&18 \\
   &  & BadT & 0&64 & 0&13 & 0&83 & 0&19 & \B{-0}&\B{01} & \B{0}&\B{21} \\
   &  & RL & 0&78 & 0&35 & 0&96 & 0&19 & 0&06 & 0&19 \\
   &  & SSD & 0&90 & 0&20 & \B{0}&\B{67} & \B{0}&\B{19} & -0&06 & 0&18 \\
   &  & LFSSD & 0&87 & 0&11 & 0&72 & 0&14 & -0&06 & 0&18 \\
\midrule
  \multirow{11}{*}{ViT} & \multirow{6}{*}{\small Full-class} & FT & 0&56 & 0&03 & 0&32 & 0&20 & 0&04 & 0&08 \\
   &  & BadT & 0&48 & 0&01 & 0&14 & 0&00 & \B{-0}&\B{00} & \B{0}&\B{00} \\
   &  & UNSIR & 0&61 & 0&03 & 0&13 & 0&01 & \B{0}&\B{00} & \B{0}&\B{00} \\
   &  & RL & 0&81 & 0&03 & \B{0}&\B{11} & \B{0}&\B{00} & 0&01 & 0&00 \\
   &  & SSD & \B{0}&\B{42} & \B{0}&\B{02} & 0&19 & 0&02 & \B{-0}&\B{00} & \B{0}&\B{00} \\
   &  & LFSSD & 0&48 & 0&01 & 0&28 & 0&05 & \B{-0}&\B{00} & \B{0}&\B{00} \\
\cmidrule{2-15}
   & \multirow{5}{*}{\small Random} & FT & \B{0}&\B{32} & \B{0}&\B{09} & 0&23 & 0&01 & \B{0}&\B{00} & \B{0}&\B{00} \\
   &  & BadT & 0&36 & 0&08 & \B{0}&\B{16} & \B{0}&\B{01} & 0&19 & 0&05 \\
   &  & RL & 1&02 & 0&08 & 0&23 & 0&01 & 0&01 & 0&04 \\
   &  & SSD & 0&63 & 0&25 & 0&61 & 0&33 & \B{0}&\B{00} & \B{0}&\B{00} \\
   &  & LFSSD & 0&84 & 0&04 & 0&87 & 0&02 & \B{0}&\B{00} & \B{0}&\B{00} \\
\bottomrule
\end{tabular}
\end{table}

\begin{table}[t]
\caption{Raw Accuracies ($\Acc_{\Dsetftest}$, $\Acc_{\Dsetrtest}$), and Membership Inference Attack ($\MIA$) values for $\Mu$ and $\Mr$ underlying the differences in Table~\ref{tab:pins_results_main} from the main paper for full-class and random unlearning on Pins Face Recognition. Accuracies are reported on the test forget split $\Dsetftest$ and the test retain split $\Dsetrtest$; $\MIA$ is computed on the training forget split $\Dsetftrain$. Full-class values are first averaged over the 5 forget classes per seed; random values are reported for the 0.1\,\% forget percentage only. All values are then reported as mean $\pm$ std across 10 seeds (260--269). UNSIR is excluded from the random scenario by design.}
\label{tab:pins_raw_values}
\centering
\resizebox{\textwidth}{!}{%
\begin{tabular}{@{}lll r@{.}l @{\,$\pm$\,} r@{.}l @{\hspace{1.5em}} r@{.}l @{\,$\pm$\,} r@{.}l @{\hspace{1.5em}} r@{.}l @{\,$\pm$\,} r@{.}l@{}}
\toprule
\textbf{Model} & \textbf{Scenario} & \textbf{Method}
& \multicolumn{4}{c}{\textbf{$\Acc_{\Dsetftest}$}}
& \multicolumn{4}{c}{\textbf{$\Acc_{\Dsetrtest}$}}
& \multicolumn{4}{c}{\textbf{$\MIA$}} \\
\midrule
  \multirow{13}{*}{ResNet18} & \multirow{7}{*}{\small Full-class} & --- reference ($\Mr$) & 0&00 & 0&00 & 97&41 & 0&12 & 0&05 & 0&01 \\
   &  & FT & 29&22 & 5&21 & 99&94 & 0&07 & 0&00 & 0&00 \\
   &  & BadT & 0&26 & 0&17 & 95&07 & 0&30 & 0&00 & 0&00 \\
   &  & UNSIR & 89&44 & 1&98 & 99&93 & 0&01 & 0&04 & 0&03 \\
   &  & RL & 0&00 & 0&00 & 100&00 & 0&00 & 0&08 & 0&02 \\
   &  & SSD & 1&97 & 6&22 & 88&28 & 7&41 & 0&06 & 0&07 \\
   &  & LFSSD & 0&00 & 0&00 & 93&76 & 1&54 & 0&02 & 0&01 \\
\cmidrule{2-15}
   & \multirow{6}{*}{\small Random} & --- reference ($\Mr$) & 88&33 & 8&47 & 97&69 & 0&22 & 0&06 & 0&18 \\
   &  & FT & 91&11 & 20&65 & 95&09 & 15&30 & 0&00 & 0&00 \\
   &  & BadT & 53&33 & 24&60 & 65&23 & 29&78 & 0&04 & 0&08 \\
   &  & RL & 39&44 & 37&45 & 93&10 & 20&16 & 0&12 & 0&13 \\
   &  & SSD & 13&33 & 21&63 & 17&73 & 23&72 & 0&00 & 0&00 \\
   &  & LFSSD & 20&00 & 15&09 & 39&09 & 19&16 & 0&00 & 0&00 \\
\midrule
  \multirow{13}{*}{ViT} & \multirow{7}{*}{\small Full-class} & --- reference ($\Mr$) & 0&00 & 0&00 & 99&78 & 0&03 & 0&00 & 0&00 \\
   &  & FT & 0&03 & 0&06 & 77&12 & 23&47 & 0&04 & 0&08 \\
   &  & BadT & 17&90 & 5&18 & 98&90 & 0&11 & 0&00 & 0&00 \\
   &  & UNSIR & 35&52 & 4&92 & 99&47 & 0&13 & 0&00 & 0&00 \\
   &  & RL & 0&00 & 0&00 & 99&98 & 0&02 & 0&02 & 0&00 \\
   &  & SSD & 0&00 & 0&00 & 97&38 & 1&00 & 0&00 & 0&00 \\
   &  & LFSSD & 0&00 & 0&00 & 95&90 & 2&09 & 0&00 & 0&00 \\
\cmidrule{2-15}
   & \multirow{6}{*}{\small Random} & --- reference ($\Mr$) & 90&00 & 5&74 & 98&58 & 0&13 & 0&00 & 0&00 \\
   &  & FT & 98&33 & 3&75 & 100&00 & 0&00 & 0&00 & 0&00 \\
   &  & BadT & 81&67 & 7&43 & 98&79 & 0&62 & 0&19 & 0&05 \\
   &  & RL & 13&33 & 8&36 & 99&91 & 0&01 & 0&01 & 0&04 \\
   &  & SSD & 35&00 & 40&83 & 44&49 & 45&61 & 0&00 & 0&00 \\
   &  & LFSSD & 0&56 & 1&76 & 7&89 & 6&59 & 0&00 & 0&00 \\
\bottomrule
\end{tabular}%
}
\end{table}

\clearpage

\section{Ethical considerations}
SUPREME is an evaluation framework for machine unlearning methods. It is method-agnostic and introduces no new unlearning method; it standardises how existing methods are compared.
The demonstration uses Pins Face Recognition, a publicly available academic benchmark of celebrity images. Identity is used as the classification target only to test whether identity-level information can be removed from a trained model. We do not propose or endorse face recognition as a deployed application. The full-class scenario uses five public-figure identities from the dataset to illustrate targeted forgetting.
SUPREME is a technical evaluation framework. Low forget-set accuracy or small parametric distance under our metrics does not constitute legal verification of data erasure under any specific regulation. 

\end{document}

%% file: supreme_diagram.tex
\begin{tikzpicture}[
    x=1in, y=1in,
    >={Stealth[length=5pt,width=4pt]},
    box/.style={draw=black!75, line width=0.7pt, rounded corners=3pt,
                fill=white, inner sep=0pt},
    stage/.style={draw=black!75, line width=0.7pt, rounded corners=3pt,
                  inner sep=0pt},
    pill/.style={draw=black!75, line width=0.4pt, rounded corners=2pt,
                 fill=black!10, inner sep=0pt,
                 minimum width=0.82in, minimum height=0.28in,
                 font=\fontsize{10}{12}\selectfont},
    arr/.style ={->, line width=0.7pt, color=black!75},
    darr/.style={->, line width=0.7pt, color=black!75, dashed},
    dline/.style={line width=0.7pt, color=black!75, dashed},
    stagetitle/.style={font=\fontsize{18}{21}\bfseries\selectfont},
    dataname/.style ={font=\fontsize{19}{22}\selectfont},
    dlname/.style   ={font=\fontsize{13}{15}\selectfont},
    devname/.style  ={font=\fontsize{12}{14}\selectfont},
    formula/.style  ={font=\fontsize{11}{13}\selectfont},
    formulasmall/.style={font=\fontsize{10}{12}\selectfont},
    synclabel/.style={font=\fontsize{12}{14}\selectfont},
    flowlabel/.style={font=\fontsize{12}{14}\selectfont},
    iterlabel/.style={font=\fontsize{13}{15}\selectfont},
    iterlabelmed/.style={font=\fontsize{13}{15}\selectfont},
    iterlabelsmall/.style={font=\fontsize{12}{14}\selectfont},
]


\draw[draw=black!60, line width=0.9pt, dash pattern=on 14pt off 6pt,
      rounded corners=4.5pt]
    (0.10, 0.55) rectangle ++(12.90, 8.75);
\node[iterlabel, anchor=north] at (6.55, 9.20)
    {For each training seed $s_{\mathrm{t}}$};

\draw[draw=black!70, line width=0.9pt, dash pattern=on 3pt off 3pt,
      rounded corners=4pt]
    (4.10, 0.74) rectangle ++(8.75, 8.11);
\node[iterlabel, anchor=north] at (8.45, 8.73)
    {For each forget target $c$};

\draw[draw=black!75, line width=0.9pt, dash pattern=on 7pt off 4pt,
      rounded corners=3.5pt]
    (4.23, 0.87) rectangle ++(8.42, 7.63);
\node[iterlabelmed, anchor=north] at (8.45, 8.43)
    {For each unlearning seed $s_{\mathrm{u}}$};

\draw[draw=black!80, line width=0.9pt, dash pattern=on 12pt off 6pt,
      rounded corners=3pt]
    (4.36, 1.70) rectangle ++(8.09, 6.40);
\node[iterlabel, anchor=north] at (8.45, 8.05)
    {For each unlearning method $a$};

\draw[draw=black!65, line width=0.6pt, dash pattern=on 2pt off 3pt,
      rounded corners=2.5pt]
    (8.43, 1.83) rectangle ++(3.84, 5.91);
\node[iterlabelsmall, anchor=north] at (10.35, 7.67)
    {For each evaluation seed $s_{\mathrm{e}}$};

\draw[stage] (0.25, 2.60) rectangle ++(3.40, 4.85);
\node[stagetitle] at (1.95, 7.05) {Stage 1: Training};

\draw[box] (1.40, 6.52) rectangle ++(1.10, 0.35);
\node[dataname] at (1.95, 6.695) {$D$};
\draw[arr] (1.95, 6.52) -- (1.95, 6.17);

\draw[box] (0.75, 5.79) rectangle ++(2.40, 0.38);
\node[dlname] at (1.95, 5.98) {DataLoader};
\draw[arr] (1.95, 5.79) -- (1.95, 5.44);

\draw[box] (0.65, 4.86) rectangle ++(2.60, 0.58);
\node[devname, anchor=north] at (1.95, 5.44) {Device 1};
\node[pill] at (1.09, 5.08) {Weights};
\node[pill] at (1.95, 5.08) {Optimiser};
\node[pill] at (2.81, 5.08) {Gradients};

\foreach \dx in {-0.10, 0.00, 0.10}
    \fill[black!45] (1.95+\dx, 4.64) circle (1.6pt);

\draw[box] (0.65, 3.86) rectangle ++(2.60, 0.58);
\node[devname, anchor=north] at (1.95, 4.44) {Device $P$};
\node[pill] at (1.09, 4.08) {Weights};
\node[pill] at (1.95, 4.08) {Optimiser};
\node[pill] at (2.81, 4.08) {Gradients};

\draw[line width=0.6pt, color=black!75] (0.85, 3.58) -- (3.05, 3.58);
\draw[line width=0.5pt] (1.00, 3.58) ++(-0.04,0.04) -- ++(0.08, 0) -- ++(-0.04,-0.08) -- cycle;
\draw[line width=0.5pt] (2.90, 3.58) ++(-0.04,0.04) -- ++(0.08, 0) -- ++(-0.04,-0.08) -- cycle;
\node[synclabel, anchor=north] at (1.95, 3.54) {Gradient Synchronisation};

\draw[box] (0.30, 2.83) rectangle ++(3.30, 0.38);
\node[formula] at (1.95, 3.02)
    {$M_{\mathrm{o}} \;\leftarrow\; \mathrm{Train}(D,\, s_{\mathrm{t}})$};

\draw[stage] (4.50, 1.95) rectangle ++(3.60, 5.50);
\node[stagetitle] at (6.30, 7.05) {Stage 2: Unlearning};

\draw[box] (5.75, 6.52) rectangle ++(1.10, 0.35);
\node[dataname] at (6.30, 6.695) {$D$};
\draw[arr] (6.30, 6.52) -- (6.30, 6.37);

\draw[box] (4.55, 5.62) rectangle ++(3.50, 0.75);
\node[formulasmall, anchor=west] at (4.68, 5.995)
    {\begin{tabular}{@{}l@{\;\;}l@{}}
     \textit{targeted:}      & $D_{\mathrm{f}} \leftarrow \{(x,y)\in D \mid y=c\}$ \\
                             & $D_{\mathrm{r}} \leftarrow D\setminus D_{\mathrm{f}}$ \\
     \textit{random-sample:} & $D_{\mathrm{f}} \leftarrow \mathrm{Sample}(D,\, c,\, s_{\mathrm{u}})$ \\
                             & $D_{\mathrm{r}} \leftarrow D\setminus D_{\mathrm{f}}$ \\
     \end{tabular}};

\draw[arr] (6.30, 5.62) -- (6.30, 5.37);

\draw[box] (5.10, 4.99) rectangle ++(2.40, 0.38);
\node[dlname] at (6.30, 5.18) {DataLoader};
\draw[arr] (6.30, 4.99) -- (6.30, 4.64);

\draw[box] (5.00, 4.06) rectangle ++(2.60, 0.58);
\node[devname, anchor=north] at (6.30, 4.64) {Device 1};
\node[pill] at (5.44, 4.28) {Weights};
\node[pill] at (6.30, 4.28) {Optimiser};
\node[pill] at (7.16, 4.28) {Gradients};

\foreach \dx in {-0.10, 0.00, 0.10}
    \fill[black!45] (6.30+\dx, 3.84) circle (1.6pt);

\draw[box] (5.00, 3.06) rectangle ++(2.60, 0.58);
\node[devname, anchor=north] at (6.30, 3.64) {Device $P$};
\node[pill] at (5.44, 3.28) {Weights};
\node[pill] at (6.30, 3.28) {Optimiser};
\node[pill] at (7.16, 3.28) {Gradients};

\draw[line width=0.6pt, color=black!75] (5.20, 2.78) -- (7.40, 2.78);
\draw[line width=0.5pt] (5.35, 2.78) ++(-0.04,0.04) -- ++(0.08, 0) -- ++(-0.04,-0.08) -- cycle;
\draw[line width=0.5pt] (7.25, 2.78) ++(-0.04,0.04) -- ++(0.08, 0) -- ++(-0.04,-0.08) -- cycle;
\node[synclabel, anchor=north] at (6.30, 2.74) {Gradient Synchronisation};

\draw[box] (4.55, 2.03) rectangle ++(3.50, 0.38);
\node[formula] at (6.30, 2.22)
    {$M_{\mathrm{u}} \;\leftarrow\; a(M_{\mathrm{o}},\, D_{\mathrm{f}},\, D_{\mathrm{r}},\, s_{\mathrm{u}})$};

\draw[box] (4.55, 1.00) rectangle ++(3.50, 0.38);
\node[formula] at (6.30, 1.19)
    {$M_{\mathrm{r}} \;\leftarrow\; \mathrm{Train}(D_{\mathrm{r}},\, s_{\mathrm{u}})$};

\draw[stage] (8.55, 1.95) rectangle ++(3.60, 5.50);
\node[stagetitle] at (10.35, 7.05) {Stage 3: Evaluation};

\draw[box] (9.15, 6.52) rectangle ++(1.10, 0.35);
\node[dataname] at (9.70, 6.695) {$D$};
\draw[box] (10.45, 6.52) rectangle ++(1.10, 0.35);
\node[dataname] at (11.00, 6.695) {$D'$};

\draw[arr] (9.70,  6.52) -- (9.70,  6.37);
\draw[arr] (11.00, 6.52) -- (11.00, 6.37);

\draw[box] (8.60, 5.62) rectangle ++(3.50, 0.75);
\node[formulasmall, anchor=west] at (8.73, 5.995)
    {\begin{tabular}{@{}l@{\;\;}l@{}}
     \textit{targeted:}      & $D'_{\mathrm{f}} \leftarrow \{(x,y)\in D' \mid y=c\}$ \\
                             & $D'_{\mathrm{r}} \leftarrow D'\setminus D'_{\mathrm{f}}$ \\
     \textit{random-sample:} & $D'_{\mathrm{f}} \leftarrow D_{\mathrm{f}}$ \\
                             & $D'_{\mathrm{r}} \leftarrow D'$ \\
     \end{tabular}};

\draw[arr] (10.35, 5.62) -- (10.35, 5.37);

\draw[box] (9.15, 4.99) rectangle ++(2.40, 0.38);
\node[dlname] at (10.35, 5.18) {DataLoader};
\draw[arr] (10.35, 4.99) -- (10.35, 4.64);

\draw[box] (9.05, 4.06) rectangle ++(2.60, 0.58);
\node[devname, anchor=north] at (10.35, 4.64) {Device 1};
\node[pill] at ( 9.92, 4.28) {$M_{\mathrm{u}}$};
\node[pill] at (10.78, 4.28) {$M_{\mathrm{r}}$};

\foreach \dx in {-0.10, 0.00, 0.10}
    \fill[black!45] (10.35+\dx, 3.84) circle (1.6pt);

\draw[box] (9.05, 3.06) rectangle ++(2.60, 0.58);
\node[devname, anchor=north] at (10.35, 3.64) {Device $P$};
\node[pill] at ( 9.92, 3.28) {$M_{\mathrm{u}}$};
\node[pill] at (10.78, 3.28) {$M_{\mathrm{r}}$};

\draw[line width=0.6pt, color=black!75] (9.25, 2.78) -- (11.45, 2.78);
\draw[line width=0.5pt] (9.40,  2.78) ++(-0.04,0.04) -- ++(0.08, 0) -- ++(-0.04,-0.08) -- cycle;
\draw[line width=0.5pt] (11.30, 2.78) ++(-0.04,0.04) -- ++(0.08, 0) -- ++(-0.04,-0.08) -- cycle;
\node[synclabel, anchor=north] at (10.35, 2.74) {All Gather};

\draw[box] (8.60, 2.03) rectangle ++(3.50, 0.38);
\node[formula] at (10.35, 2.22)
    {$R \;\leftarrow\; \mathrm{Evaluate}(M_{\mathrm{u}},\, M_{\mathrm{r}},\, D'_{\mathrm{f}},\, D'_{\mathrm{r}},\, E,\, s_{\mathrm{e}})$};


\draw[dline] (1.95, 2.60) -- (1.95, 2.30);
\draw[dline] (1.95, 2.30) -- (3.80, 2.30);
\draw[dline] (3.80, 2.30) -- (3.80, 6.95);
\draw[dline] (3.80, 6.95) -- (3.90, 6.95);
\draw[darr]  (3.90, 6.95) -- (4.10, 6.95);

\node[flowlabel] at (2.875, 2.12) {Trained model $M_{\mathrm{o}}$};

\draw[dline] (8.05, 2.22) -- (8.30, 2.22);   
\draw[dline] (8.05, 1.19) -- (8.30, 1.19);   
\draw[dline] (8.30, 1.19) -- (8.30, 6.95);   
\draw[darr]  (8.30, 6.95) -- (8.43, 6.95);   

\end{tikzpicture}